\documentclass[runningheads]{llncs}

\usepackage{eccv}

\usepackage{eccvabbrv}

\usepackage{graphicx}
\usepackage{booktabs}

\usepackage[accsupp]{axessibility}  

\usepackage{algorithm}
\usepackage{algorithmic}

\usepackage{amsmath}
\usepackage{epsfig}
\usepackage{enumerate}
\usepackage{float}
\usepackage{color}
\usepackage{amssymb}
\usepackage{tabularx}
\usepackage{multirow}
\usepackage{subcaption}
\usepackage{indentfirst}
\usepackage{booktabs}
\usepackage{adjustbox}
\usepackage{tabularray}
\usepackage[table,xcdraw]{xcolor}
\usepackage{wrapfig}
\usepackage{bbding}

\usepackage[colorlinks=true, citecolor=eccvblue, linkcolor=eccvblue, urlcolor=eccvblue]{hyperref}

\usepackage{orcidlink}

\newcolumntype{x}[1]{>{\centering\arraybackslash}p{#1pt}}
\newcolumntype{y}[1]{>{\raggedright\arraybackslash}p{#1pt}}
\newcolumntype{z}[1]{>{\raggedleft\arraybackslash}p{#1pt}}

\newcommand{\increase}[1]{
	\fontsize{0.3pt}{0.1em}\selectfont\color{purple}{$\uparrow$~\textbf{#1}}
}
\newcommand{\decrease}[1]{
	\fontsize{0.3pt}{0.1em}\selectfont\color{gray!48}{$\downarrow$~\textbf{#1}}
}
\newcommand{\nochange}[1]{
    \fontsize{0.3pt}{0.1em}\selectfont\color{gray}{--~\textbf{#1}}
}

\newcommand{\xmltag}[1]{\texttt{\char`\<#1\char`\>}}

\definecolor{cred}{HTML}{FF6B6B}
\definecolor{cyellow}{HTML}{FEC260}
\definecolor{cgreen}{HTML}{70AD47}
\definecolor{cblue}{HTML}{4D96FF}
\definecolor{cpurple}{HTML}{2A0944}
\definecolor{ggray}{RGB}{127,127,127}
\definecolor{aliceblue}{rgb}{0.94, 0.97, 1.0}
\definecolor{deepgreen}{RGB}{0,128,0}

\begin{document}

\title{GAIA: A Data Flywheel System for Training GUI Test-Time Scaling Critic Models} 

\titlerunning{A Data Flywheel System for Training GUI Test-Time Scaling Critic Models}

\author{Shaokang Wang~\thanks{This paper was completed during the internship at Xiaomi. \Envelope~Corresponding author.}~\inst{1,2} \and
Pei Fu\inst{1} \and
Ruoceng Zhang\inst{1} \and
Shaojie Zhang\inst{1} \and
Xiuwen Xi\inst{1} \and
Jiahui Yang\inst{1} \and
Bin Qin\inst{1} \and
Ying Huang\inst{1} \and
Zhenbo Luo~\Envelope~\inst{1} \and
Jian Luan\inst{1}}

\authorrunning{S. Wang et al.}

\institute{MiLM Plus, Xiaomi Inc, Beijing, China \and
School of Computer Science, Peking University, Beijing, China \\
\email{skwang6272@stu.pku.edu.cn, \{fupei1, luozhenbo\}@xiaomi.com}}

\maketitle

\begin{abstract}
While Large Vision-Language Models (LVLMs) have significantly advanced GUI agents' capabilities in parsing textual instructions, interpreting screen content, and executing tasks, a critical challenge persists: the irreversibility of agent operations—where a single erroneous action can trigger catastrophic deviations. 
To address this, we propose the \textbf{G}UI \textbf{A}ction Cr\textbf{i}tic's Dat\textbf{a} Flywheel System (GAIA), a training framework that enables the models to have iterative critic capabilities, which are used to improve the Test-Time Scaling (TTS) of basic GUI agents' performance.
Specifically, we train an \textbf{Intuitive Critic Model} (ICM) using positive and negative action examples from a base agent first. This critic evaluates the immediate correctness of the agent's intended actions, thereby selecting operations with higher success probability.
Then, the initial critic guides agent actions to collect refined positive/negative samples, initiating the self-improving cycle. The augmented data then trains a second-round critic with enhanced discernment capability.
We conduct experiments on various datasets and demonstrate that the proposed ICM can improve the test-time performance of various closed-source and open-source models, and the performance can be gradually improved as the data is recycled. The code, dataset, and accompanying datasheet will be publicly released at \url{https://github.com/SeerRay-Lab/GAIA}.
\end{abstract}

\section{Introduction}
\label{sec:intro}
The automation of Graphical User Interface~(GUI) interactions represents a critical frontier in developing intelligent digital assistants~\cite{wang2024gui,hu2024agents,nguyen2024gui}. Recent breakthroughs in Large Vision-Language Models (LVLMs)~\cite{wang2024qwen2,bai2025qwen2}, leveraging advanced post-training techniques, have substantially enhanced agents' capabilities in interpreting natural language commands, perceiving visual elements, and executing multi-step tasks~\cite{hong2024cogagent,cheng2024seeclick}.
Within this rapidly evolving landscape, the development of robust GUI agents has largely converged on two primary methodological paradigms. The first approaches \cite{wu2024atlas,xu2024aguvis,qin2025ui,liu2025infiguiagent} train models through Supervised Fine-Tuning~(SFT) to directly align their behavior with predefined task objectives. The second approaches employ Reinforcement Fine-Tuning~(RFT)~\cite{lu2025ui,xia2025gui,liu2025infigui}, which significantly enhances generalization in complex tasks by adopting a reasoning format.

Despite these advances, the dynamic and continuous nature of real-world GUI tasks means that agents can still produce ambiguous or incorrect action proposals at any step. \textbf{A single mis-click or mis-typed output can be irreversible}, derailing the entire workflow and leaving the system in an unrecoverable state. This high-stakes environment imperatively demands \textbf{a mechanism for pre-execution validation.}

\begin{figure}[t]
    \centering
    \begin{subfigure}[t]{\columnwidth}
        \centering
        \includegraphics[width=0.9\linewidth]{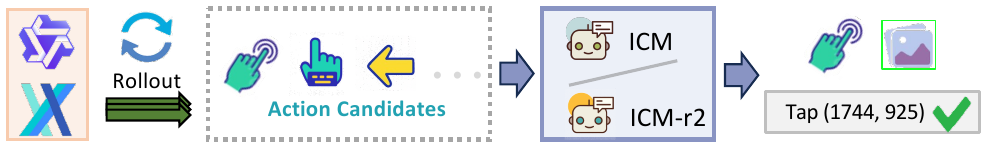}
        \caption{The promotion process of the critic model to the GUI Agent during testing.}
        \label{fig:fig1_1}
    \end{subfigure}
    

    \begin{subfigure}[t]{\columnwidth}
        \centering
        \begin{minipage}[t]{0.43\linewidth}
            \centering
            \includegraphics[width=\linewidth]{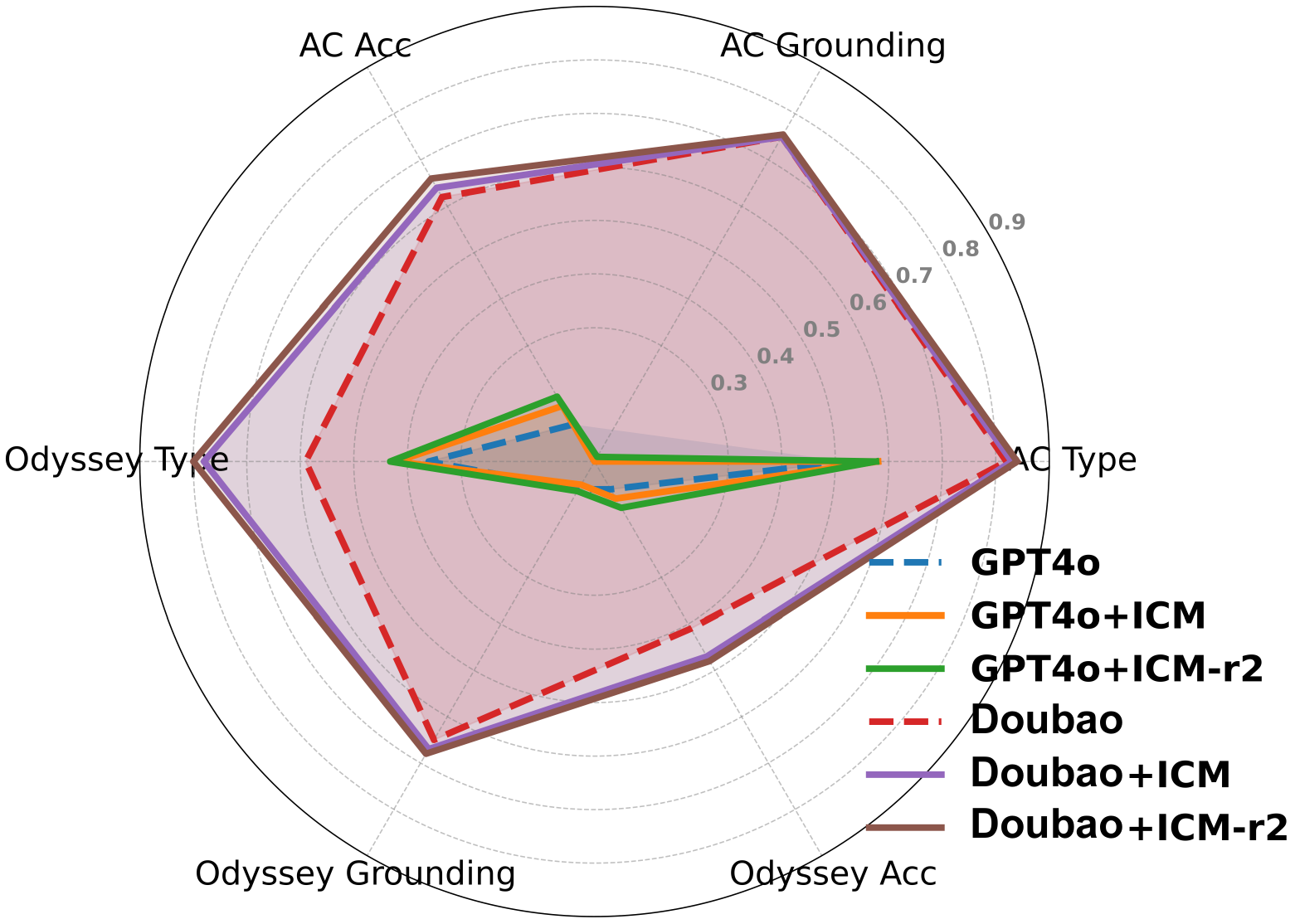}
        \end{minipage}
        \begin{minipage}[t]{0.43\linewidth}
            \centering
            \includegraphics[width=\linewidth]{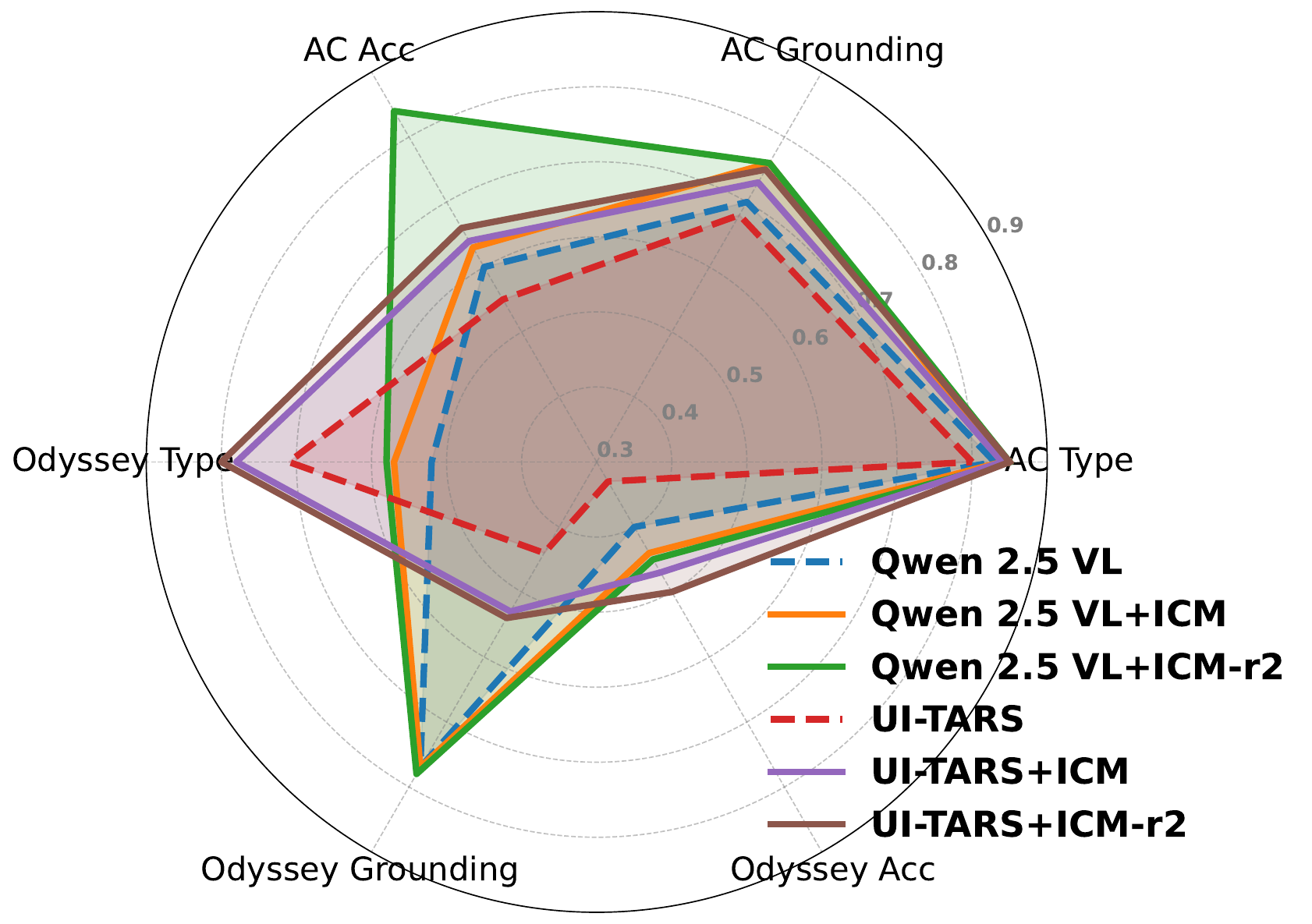}
        \end{minipage}
        \caption{Comparison of high-level task performance improvements on closed-source and open-source GUI Agents.}
        \label{fig:comparison}
    \end{subfigure}
    \caption{\textbf{Intuitive results.} ICM guides agents' action during testing, as shown in (a), thereby improving the agent's accuracy, and is continuously improved by the data flywheel, as shown in (b).}
    \label{fig_intuitive}
\end{figure}

To avoid irreversible errors in execution and improve the performance of basic GUI agents during testing, previous studies have designed action verifiers for GUI agents~\cite{wu2025gui,xiao2025ui,yang2025gta1}, which are used to judge and filter GUI agents' actions. However, these existing implementations suffer from two primary limitations.
First, training a correctness verifier requires defining positive and negative action samples. Existing work on defining negative samples relies on heuristic algorithms, such as randomly selecting click locations on the current screenshot~\cite{xiao2025ui}, which fails to capture the realistic action distribution and leads to suboptimal judgment performance.
Second, the reasoning-based verifiers~\cite{wanyan2025look} implemented in existing work violate the intuitive properties of binary judgments. For an intuitive correctness judgment problem, biological research suggests that higher-level judgment pathways are often more adept than performing extensive multi-step reasoning~\cite{liu2011neural,poldrack2005neural,doyon2005reorganization}, which indicates that excessive reasoning can be less effective~\cite{bilalic2008good,wan2011neural}. Furthermore, reasoning-based judgment outputs more tokens, thereby reducing the efficiency of test-time scaling.

To fully leverage pre-execution evaluation to enhance GUI agent capabilities and execution correctness, we developed a \textbf{G}UI \textbf{A}ction Cr\textbf{i}tic's Dat\textbf{a} Flywheel System (GAIA). This system comprises two core phases: the initialization phase (Phase 1) and the iteration phase (Phase 2), yielding the \textbf{Intuitive Critic Model} (ICM).
In \textbf{Phase 1}, we use real GUI agents to act on an existing dataset to collect positive and negative action data that are random but consistent with the behavior distribution. Using this binary-labeled action dataset, we train ICM to assess action correctness given environmental context.
In \textbf{Phase 2}, as illustrated in Figure \ref{fig_intuitive}(a), ICM employs a Best-of-N approach to select the highest-probability correct actions from agent rollouts. While ICM guidance significantly improves action accuracy, challenging samples persist and produce errors. These difficult cases are annotated and fed back into the data flywheel. Through iterative data augmentation, the flywheel continuously incorporates new action samples, progressively covering challenging scenarios within the action space. Driven by this enriched dataset, we train an enhanced critic—Intuitive Critic Model on Round Two (ICM-r2)—which achieves higher discriminative accuracy for more precise behavioral guidance. This establishes a self-evolutionary virtuous cycle between the data flywheel and critic models, continuously improving GUI agent action accuracy.

Leveraging the proposed system GAIA, ICM achieves SOTA performance in action critique. Naturally, we integrate it into Test-Time Scaling (TTS)~\cite{snell2025scaling,chen2024expanding,snell2024scaling,prabhudesai2023diffusion,wang2025mcts,tian2025think} during inference, where ICM evaluates stochastically generated actions from the TTS process, releasing only high-confidence operations surpassing predetermined thresholds for execution. To validate the framework's general applicability, we conduct joint experiments using mainstream GUI Agents (including GPT-4o~\cite{hurst2024gpt} and UI-TARS~\cite{qin2025ui}) on several GUI  Agent benchmarks.
As shown by the comparative results in Figure~\ref{fig_intuitive}~(b), the guidance from our iteratively evolved critic models (ICM and ICM-r2) leads to significant performance improvements in basic GUI agents, including GUI operation task planning and grounding capabilities.

Overall, the main contributions are summarized as follows:
\begin{enumerate}
    \item We introduce GAIA—a novel Data Flywheel System designed for training GUI action-critic models. By iteratively curating positive and negative samples from real-world action data, GAIA continuously boosts model performance and robustness.
    \item We propose the ICM for GUI interaction tasks, a critic model trained on data curated by our data flywheel. The ICM enhances the performance of existing GUI agents by employing a best-of-N approach to select the most probable correct action with TTS. This initial boost is then continuously refined as the ICM's discriminatory accuracy is iteratively improved by the data flywheel.
    \item We comprehensively demonstrate across multiple datasets that ICM trained with our proposed GAIA system significantly enhances the overall performance of both closed-source and open-source GUI agents.
\end{enumerate}

\section{Related Work}
\subsection{GUI Agent}
The development of autonomous agents powered by LLMs and LVLMs has significantly advanced interactive functionalities within digital environments. 
Early GUI systems primarily leveraged LLMs to interpret structured representations~\cite{hong2024cogagent,nong2024mobileflow,song2024visiontasker}.
The development of LVLM simplifies the paradigm, allowing GUI agents to receive raw visual signals from the screenshots~\cite{hu2024agents,liu2024autoglm,shen2024falcon,tang2025think,christianos2024lightweight,zheng2025vem,gou2024navigating,wu2024atlas}. 
Recent efforts, such as Aguvis~\cite{xu2024aguvis} and UI-TARS~\cite{qin2025ui}, have advanced autonomous GUI navigation by integrating explicit planning, sophisticated reasoning, and GUI-specific pretraining to handle complex digital environments.
Concurrently, the advent of rule-based Reinforcement Learning~(RL) approaches~\cite{jaech2024openai,guo2025deepseek} has further enhanced GUI agent capabilities. These RFT methods improve reasoning and generalization by enabling models to learn universal action strategies from high-quality samples~\cite{liu2025visual,shen2025vlm,lu2025ui,xia2025gui,liu2025infigui}. 
While fine-tuning and model scaling can enhance GUI agent capabilities, these methods are often prohibitively resource-intensive. This highlights a clear need for test-time enhancements that can offer universal performance improvements across various agent models without costly retraining.

\begin{figure*}[t]
    \centering
    \includegraphics[width=\textwidth]{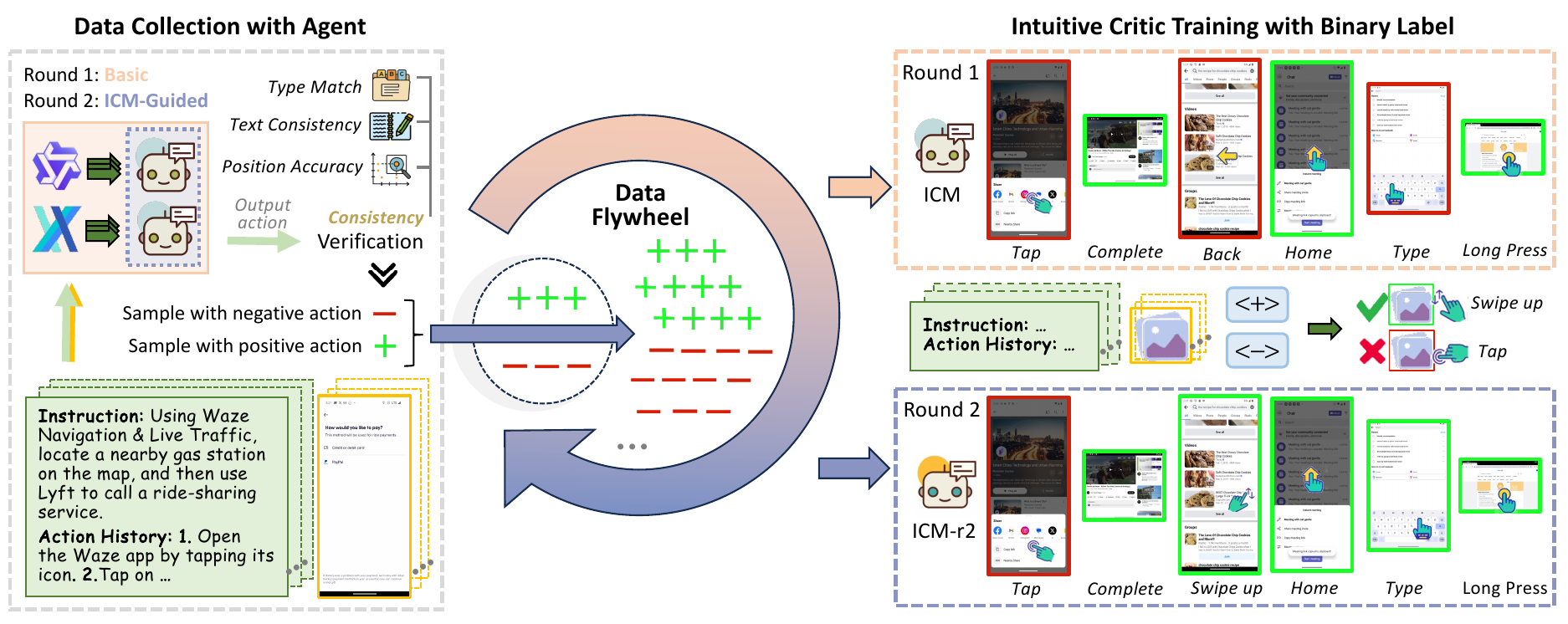}
    \caption{\textbf{Data flywheel curation pipeline for GAIA.} A sample dataset is constructed using GUI agent interactions. The positive and negative labels are marked by comparing the ground truth actions to train an action correctness discrimination model. After the critic model guides the GUI agent, it further expands the dataset, pushing the data flywheel to cover more action distributions, thereby promoting the iterative improvement of model performance.}
    \label{fig2}
\end{figure*}

\subsection{Critic Model}
To solve the problem of suboptimal single-shot model output~\cite{zhang2025llm,martino2023knowledge,wen2024reinforcing,chen2024optima}, research has gradually focused on improving the performance of the basic model during testing with the help of the critic model~\cite{mcaleese2024llm,ji2023towards,kalyanpur2024llm,zhang2025critic,xiong2025llava}. This concept has been expanded to the GUI domain with notable works like GUI-Genie~\cite{xiao2025ui}, GUI-Actor~\cite{wu2025gui}, GTA1~\cite{yang2025gta1}, and GUI-Critic-R1~\cite{wanyan2025look}.
However, existing GUI critics often rely on synthetic data generated by heuristic algorithms, such as randomly selecting click locations~\cite{wu2025gui}, cross-task substitution, or early truncation~\cite{xiao2025ui}. This approach fails to accurately simulate the complex behavior of real GUI agents across the full action space, thereby preventing the critic from learning faithful discrimination criteria. 
Furthermore, while some approaches use RL to inject reasoning capabilities into the critic~\cite{wanyan2025look}, this often contradicts the very motivation for intuitive judgment~\cite{liu2011neural,wan2011neural} and introduces delays due to extended output token generation.

\section{Method}
In this section, we detail the design of our data flywheel-driven GAIA system for the GUI agent shown in Figure~\ref{fig2}. We begin in Section 3.1 by introducing the general definition of the GUI agent task and the crucial role of the critic model. Section 3.2 delves into the design and application of our data flywheel system within the initial round of the evaluation process. In Section 3.3, we present the model training in the second round, which builds upon the outcomes from the first iteration and forms a virtuous cycle.

\subsection{Preliminaries}
\label{method_pre}
The interaction between a GUI agent and its environment can be formulated as a Markov Decision Process~(MDP), denoted by the tuple $\langle \mathcal{S}, \mathcal{A}, \mathcal{Z}, \mathcal{T}, \mathcal{O} \rangle$. Here, $\mathcal{S}$ defines the state space of possible screen states, while $\mathcal{A}$ encompasses the action space, including interaction types like clicking, typing, and scrolling. The observation space $\mathcal{Z}$ captures inputs such as screenshots or structured UI representations. The state transition probability is given by $\mathcal{T}: \mathcal{S} \times \mathcal{A} \times \mathcal{S} \rightarrow [0,1]$, mapping a state and action to a new state. Similarly, $\mathcal{O}: \mathcal{S} \times \mathcal{A} \rightarrow \mathcal{Z}$ describes the likelihood of observing a particular output given a state and an action. During GUI task execution, at each discrete time step $t$, the agent receives an input tuple $(z_t, u, h)$, comprising the current screen observation $z_t \in \mathcal{Z}$, the global task instruction $u$, and the accumulated interaction history $h$. The agent's decision-making process for GUI actions is then formalized by a structured policy function $\mathcal{F}$:
\begin{equation}
    \mathcal{F}(z_t,u,h) \rightarrow o_t = \{a_t, c_t\},
\end{equation}
where $o_t$ represents the agent output at time $t$, consisting of the action type $a_t$ (e.g., \emph{click}, \emph{swipe}, and \emph{type}) and its corresponding parameters $c_t$ (e.g., click coordinates, text content for typing). After $a_t$ is executed, the environment transitions to a new state $z_{t+1}$, and this iterative process continues until the task is successfully completed or a predefined termination condition is met.

\begin{figure*}[t]
    \centering
    \includegraphics[width=\textwidth]{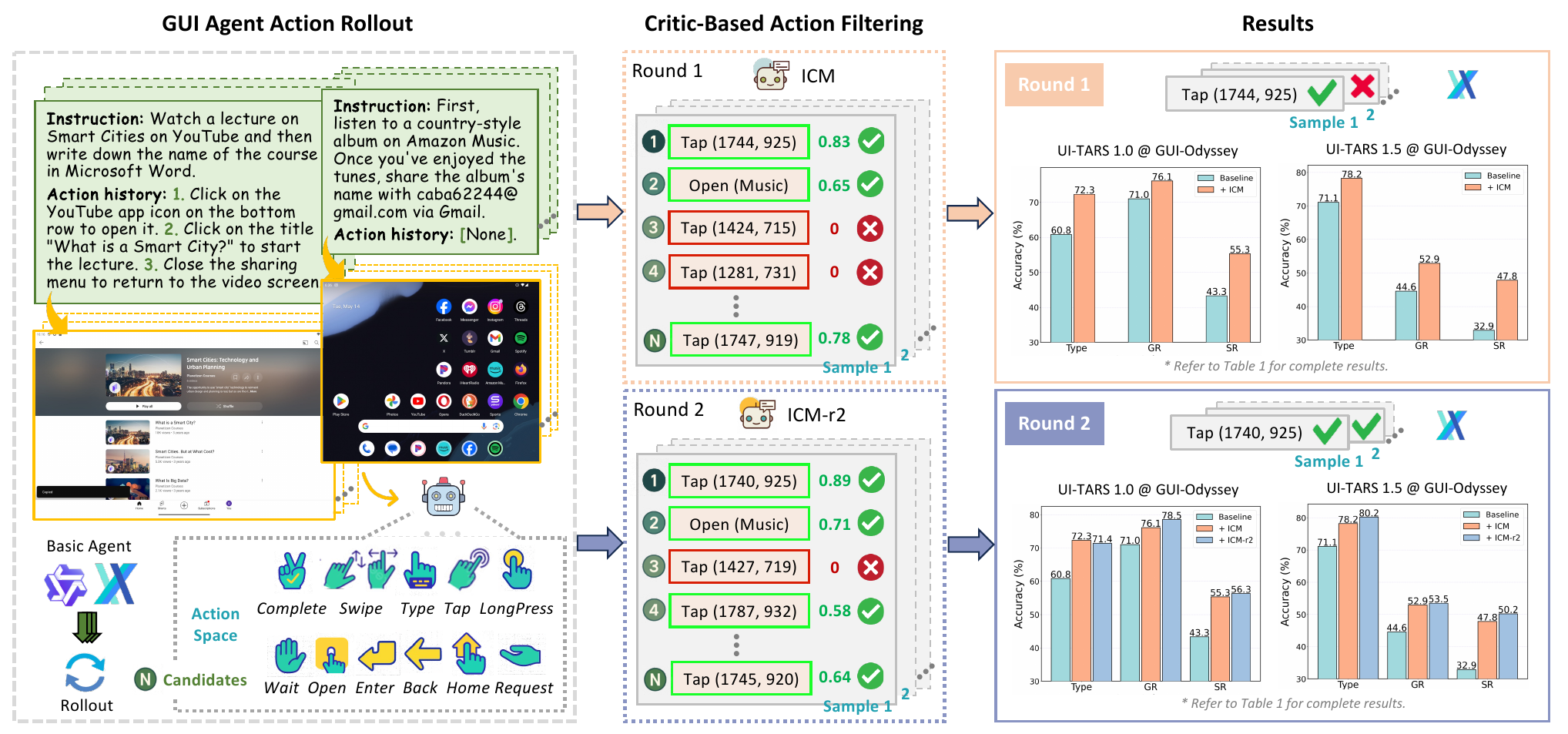}
    \caption{\textbf{Test-Time scaling pipeline.} Through best-of-N rollout, multi-candidate actions of GUI agents are given, and the correct action with the highest probability is selected after ICM evaluation.}
    \label{fig3}
\end{figure*}

The proposed ICM, building upon the same observations and the GUI agent's current proposed action $o_t$, outputs a judgment $j_t$ regarding the correctness of that action:
\begin{equation}
    \mathcal{J}(o_t | (z_t,u,h)) \rightarrow l_t = \{j_t, p_t\},
\end{equation}
where $j_t$ is a binary indicator, ``\textit{correct}" for correct actions and ``\textit{wrong}" for incorrect, $p_t$ represents the probability of the judgment, which supports finding the correct action with the highest confidence. By enabling the sampling of multiple candidate actions and prioritizing them based on their respective correctness probabilities, ICM ensures that a more optimal action for the current state is selected and executed, significantly enhancing the agent's actual success rate.

\subsection{Action Decision with Intuitive Critic}

\subsubsection{Data Curation}
To enable the judgment model to distinguish the correctness of real actions, we meticulously define both positive and negative samples of GUI agent actions. We begin by having existing GUI agents $\Pi = \{\pi_1, \pi_2, ..., \pi_i\}$~\cite{qin2025ui} interact with and traverse publicly accessible datasets~\cite{li2024effects,lu2024gui}, allowing us to collect authentic, step-level operations across various GUI scenarios. For each action executed in a specific state $(z,u,h)$, we then leverage ground truth labels to determine its correctness. An action is designated positive (with a correctness judge $j=``\textit{correct}"$) if it aligns with the GT.

Conversely, we identify negative samples (with a correctness score $j=``\textit{wrong}"$) based on states where the agent's action deviates from the GT. This approach ensures that our collected negative operations are closely aligned with the actual error distribution observed in real GUI environments, significantly enhancing the quality and realism of our training dataset. To prevent bias during ICM training, we balance the collected positive and negative samples, ensuring an equal 50\% split for each. This carefully curated dataset, denoted as $\mathcal{D} = \{j_k | z_k,u_k,h_k,o_k^{\pi_i}\}_{k=1}^K$, forms the foundation of our data flywheel GAIA.

\subsubsection{ICM Training and Guidance}
Based on the dataset $\mathcal{D}$, ICM is trained to intuitively judge the correctness of actions. Specifically, the input of ICM includes the screen observation $ z_k$, the instruction description $u_k$, the action history $h_k$, and the given agent action $o_k^{\pi_i}$. We implement ICM using LVLM and use standard cross-entropy loss to supervise the ICM's output tokens. For each sample in our dataset D, the model's output is a token representing either ``\textit{correct}" or ``\textit{wrong}". The training process aims to minimize the discrepancy between the model's predicted probability and the ground truth label:
\begin{equation}
\begin{split}
& \mathcal{L}_{\text{CE}} = -\frac{1}{K} \sum_{k=1}^{K} \biggl[ j_k \log\left(P_{\theta_c}(\textit{``correct''} \mid z_k, u_k, h_k, o_k^{\pi_i})\right) \\
  + & (1 - j_k) \log\left(1 - P_{\theta_c}(\textit{``correct''} \mid z_k, u_k, h_k, o_k^{\pi_i})\right) \biggr],
\end{split}
\label{eq_loss}
\end{equation}
where $P_{\theta_c}(\textit{``correct''} \mid z_k, u_k, h_k, o_k^{\pi_i})$ represents the probability assigned by the critic model $\theta_c$ to the ``\textit{correct}" token.

During test-time, a GUI agent $\pi_i$ generates $N$ candidate actions $\mathcal{O} = \{o_1, \dots, \allowbreak o_N\}$ through N-rollout sampling. ICM evaluates these candidates by assigning each action a correctness judge $j_n$ and a confidence score $s_n$. Leveraging the best-of-N filtering strategy, we select the optimal action $o^*$ from the subset of correct candidates $\mathcal{O}_{\text{correct}}$ that has the highest confidence score, which is formalized as:
\begin{equation}
    o^* = 
    \begin{cases}
    \arg\max\limits_{o_n \in \mathcal{O}_{\text{correct}}} s_n, & \text{if } \mathcal{O}_{\text{correct}} \neq \varnothing \\
    o_1. & \text{otherwise}
    \end{cases}
    \label{eq_guide}
\end{equation}
This approach effectively guides the agent to bypass single-shot output failures and select the most promising action, thereby significantly boosting its overall execution accuracy. 

\begin{table*}[!t]
    \renewcommand{\arraystretch}{1.15} 
    \small
    \centering
    \caption{\textbf{Data distribution of the flywheel.} $\mathcal{D}$ and $\mathcal{D}^+$ respectively represent the data of the first and the second round of GAIA.}
    \label{tab_data}
    \begin{tabular}{y{40}y{80}y{80}y{80}}
        \noalign{\hrule height 0.8pt}
        \hline
        Category & Source Dataset & Positive Sample & Negative Sample\\
        \hline
        \multirow{2}{*}{$\mathcal{D}$} & AndroidControl & 68.2k & 69.9k \\
        & GUI-Odyssey & 65.4k & 66.8k \\
        \hline
        \multirow{2}{*}{$\mathcal{D}^+$} & AndroidControl & (68.2+15.1)k & (69.9+14.0)k \\
        & GUI-Odyssey & (65.4+26.1)k & (66.8+26.3)k \\
        \noalign{\hrule height 1.0pt}
    \end{tabular}
\end{table*}

\subsection{Data Flywheel and Critic Scaling}
Guided by Equation~\ref{eq_guide}, the execution accuracy has been significantly improved. However, some difficult action samples require more precise judgment. Considering that ICM and test-time scaling performance can be further enhanced with data, we collect agent actions guided by ICM and, after filtering for positive and negative balance, add them to the data flywheel to form $\mathcal{D}^+ = \{j_k | z_k,u_k,h_k,o_k^{\pi_i},\theta_c\}_{k=1}^{K'}$. $\mathcal{D}^+$ further covers the distribution of actions, providing a foundation for performance scaling.

Based on the challenging samples in this enriched dataset, we train the ICM on Round Two~(ICM-r2), using the same cross-entropy loss as defined in Equation~\ref{eq_loss}. This new dataset, which is specifically curated to expose the critic's most significant blind spots, allows ICM-r2 to acquire a more nuanced and accurate discriminative ability. Consequently, as illustrated in Figure~\ref{fig3}, ICM-r2 provides more precise guidance for the agent's action selection, thereby fundamentally strengthening the critic's overall judgment and significantly improving the agent's performance on the most difficult tasks. Together with ICM, ICM-r2 demonstrates the power of a data flywheel-driven approach to stimulate the performance of GUI agents during testing.

\section{Experiment}

\subsection{Implementation Details}
\label{exp_implementation}

\textbf{Experimental Setup.} 
We use UI-TARS 1.0~\cite{qin2025ui} and UI-TARS 1.5~\cite{qin2025ui} for inference on the AndroidControl~\cite{li2024effects} and GUI-Odyssey~\cite{lu2024gui} training sets, and compare the real actions with GT to build $\mathcal{D}$ and $\mathcal{D}^+$.
On the corresponding data, we develop the ICM and ICM-r2 based on Qwen2.5 VL 7B~\cite{bai2025qwen2} and adopt the ms-swift~\cite{zhao2024swiftascalablelightweightinfrastructure} framework for training.
All action judgments followed the high-level approach, providing only global instructions to the ICM and ICM-r2, not single-step instructions. The distribution of the data flywheel is shown in Table~\ref{tab_data}. The critic model guides the agents in the N-rollout process with $N=8$. To allow the base agent to sample a reasonable range of potential actions, its temperature coefficient, top\_k, and top\_p are set to 1.0, 30, and 0.8, respectively. All experiments are conducted on 8 NVIDIA H100-80G GPUs.

\textbf{Action Space.}
We define a unified action space including \emph{Click}, \emph{Swipe}, \emph{Type}, \emph{Open}, \emph{Home}, \emph{Back}, \emph{Enter}, and \emph{Wait}. Parameterized actions (\emph{Click}, \emph{Swipe}, \emph{Type}, \emph{Open}) are associated with structured arguments such as coordinates, directions, text content, or app identifiers.

Binary critic labels are obtained by comparing normalized agent actions with benchmark ground truth following dataset protocols~\cite{li2024effects,lu2024gui}. Specifically, \emph{Click} actions have a coordinate error within a 14\% screen-distance threshold to the ground truth, \emph{Swipe} matches direction, \emph{Type} and \emph{Open} require normalized text or app-name matching, and system actions (\emph{Home}, \emph{Back}, \emph{Enter}, \emph{Wait}) are matched according to discrete states.

To better understand the distribution and characteristics of the collected data, we sample 300 auto-labeled negatives from $\mathcal{D}$ for analysis. \emph{Click} accounts for 64\%, followed by \emph{Open} (13\%), \emph{Swipe} (9\%), and \emph{Type} (4\%), with the remaining actions comprising the rest. Detailed dataset documentation is provided in the supplementary material.

\begin{table}[!t]
    \renewcommand{\arraystretch}{1.15}
    \scriptsize
    \centering
    \caption{\textbf{GUI planning accuracy on AndroidControl.} $^\dag$ represents the closed-source UI-TARS 1.5. $^*$ represents a reproduced agent. $\color{purple}{\uparrow}$ and $\color{gray!48}\downarrow$ represent changes relative to base agents.}
    \label{tab_plan_android}
    \begin{tabular}{y{55}y{20}x{32}y{33}y{33}y{33}y{33}y{36}y{33}}
        \noalign{\hrule height 1.0pt}
        \multirow{2}{*}{Model} & \multirow{2}{*}{Method} & \multirow{2}{*}{Size} & \multicolumn{3}{c}{AndroidControl-Low} & \multicolumn{3}{c}{AndroidControl-High} \\ \cmidrule(lr){4-6} \cmidrule(lr){7-9}
                             & & & Type & GR & SR & Type & GR & SR \\ \hline
        OS-Atlas-Base & ZS & 7B & 73.0 & 73.4 & 50.9 & 57.4 & 54.9 & 29.8 \\
        SeeClick & SFT & 9.6B & 93.0 & 73.4 & 75.0 & 82.9 & 62.9 & 59.1 \\
        Aria-UI & SFT & 3.9B & – & 87.7 & 67.3 & – & 43.2 & 10.2 \\
        Aguvis & SFT & 7B & – & – & 80.5  & – & – & 61.5 \\
        UI-R1 & RFT & 3B & 79.2 & 82.4 & 66.4 & 57.9 & 55.7 & 45.4 \\
        GUI-R1 & RFT & 3B & 83.7 & 81.6 & 64.4 & 58.0 & 56.2 & 46.6 \\ \hline
        \multicolumn{2}{l}{GPT4o} & – & 78.8 & 8.0 & 20.4 & 52.4 & 3.6 & 13.2 \\ 
        \multicolumn{2}{r}{+ ICM} &  & 82.4~\increase{3.6} & 9.2~\increase{1.2} & 24.8~\increase{4.4} & 58.0~\increase{5.6} & 5.3~\increase{1.7} & 17.0~\increase{3.8} \\
        \rowcolor{aliceblue!100} \multicolumn{2}{r}{+ ICM-r2} &  & 81.6~\increase{2.8} & 8.5~\increase{0.5} & 23.8~\increase{3.4} & 57.6~\increase{5.2} & 6.1~\increase{2.5} & 18.8~\increase{5.6} \\
        \multicolumn{2}{l}{$\text{Doubao}^\dag$} & – & 97.0 & 86.4 & 86.2 & 82.0 & 75.0 & 62.4 \\ 
        \multicolumn{2}{r}{+ ICM} & & 97.0~\nochange{} & 86.6~\increase{0.2} & 86.4~\increase{0.2} & 83.2~\increase{1.2} & 75.1~\increase{0.1} & 64.4~\increase{2.0} \\
        \rowcolor{aliceblue!100} \multicolumn{2}{r}{+ ICM-r2} & & 96.6~\decrease{0.4} & 86.6~\increase{0.2} & 86.0~\decrease{0.2} & 84.2~\increase{2.2} & 75.5~\increase{0.5} & 66.0~\increase{3.6} \\
        \multicolumn{2}{l}{Qwen 2.5 VL} & 7B & 94.4 & 85.6 & 81.8 & 83.0 & 70.9 & 60.6 \\ 
        \multicolumn{2}{r}{+ ICM} & & 95.4~\increase{1.0} & 86.0~\increase{0.4} & 81.2~\decrease{0.6} & 84.0~\increase{1.0} & 75.9~\increase{5.0} & 63.4~\increase{2.8} \\
        \rowcolor{aliceblue!100} \multicolumn{2}{r}{+ ICM-r2} & & 94.6~\increase{0.2} & 85.4~\decrease{0.2} & 81.8~\nochange{} & 84.4~\increase{1.4} & 75.9~\increase{5.0} & 63.8~\increase{3.2} \\
        \multicolumn{2}{l}{$\text{UI-TARS 1.0}^*$} & 7B & 90.0 & 85.1 & 75.4 & 80.8 & 68.9 & 58.2 \\ 
        \multicolumn{2}{r}{+ ICM} & & 90.5~\increase{0.5} & 87.6~\increase{2.5} & 80.4~\increase{5.0} & 82.3~\increase{1.5} & 79.5~\increase{10.6} & 67.5~\increase{9.3} \\
        \rowcolor{aliceblue!100} \multicolumn{2}{r}{+ ICM-r2} & & 90.0~\nochange{} & 86.8~\increase{1.7} & 80.2~\increase{4.8} & 82.7~\increase{1.9} & 78.9~\increase{10.0} & 67.1~\increase{8.9} \\
        \multicolumn{2}{l}{$\text{UI-TARS 1.5}^*$} & 7B & 86.4 & 82.4 & 72.2 & 80.2 & 68.1 & 55.8 \\ 
        \multicolumn{2}{r}{+ ICM} & & 90.3~\increase{3.9} & 85.0~\increase{2.6} & 79.0~\increase{6.8} & 84.2~\increase{4.0} & 73.3~\increase{5.2} & 64.5~\increase{8.7} \\
        \rowcolor{aliceblue!100} \multicolumn{2}{r}{+ ICM-r2} & & 90.1~\increase{3.7} & 85.5~\increase{3.1} & 79.2~\increase{7.0} & 84.6~\increase{4.4} & 74.7~\increase{6.6} & 65.6~\increase{9.8} \\
        \noalign{\hrule height 1.0pt}
    \end{tabular}
\end{table}

\begin{table}[!t]
    \renewcommand{\arraystretch}{1.15}
    \scriptsize
    \centering
    \caption{\textbf{GUI planning accuracy on GUI-Odyssey.} $^\dag$ represents the closed-source UI-TARS 1.5. $^*$ represents a reproduced agent. $\color{purple}{\uparrow}$ and $\color{gray!48}\downarrow$ represent changes relative to base agents.}
    \label{tab_plan_odyssey}
    \begin{tabular}{y{60}y{30}x{35}y{43}y{43}y{43}}
        \noalign{\hrule height 1.0pt}
        \multirow{2}{*}{Model} & \multirow{2}{*}{Method} & \multirow{2}{*}{Size} & \multicolumn{3}{c}{GUI-Odyssey} \\ \cmidrule(lr){4-6}
                             & & & Type & GR & SR \\ \hline
        OS-Atlas-Base & ZS & 7B & 60.4 & 39.7 & 27.0 \\
        SeeClick & SFT & 9.6B & 71.0 & 52.4 & 53.9 \\
        Aria-UI & SFT & 3.9B & – & 86.8 & 36.5 \\
        Aguvis & SFT & 7B & – & – & – \\
        UI-R1 & RFT & 3B & 52.2 & 34.5 & 32.5 \\
        GUI-R1 & RFT & 3B & 54.8 & 41.5 & 41.3 \\ \hline
        \multicolumn{2}{l}{GPT4o} & – & 36.6 & 11.6 & 11.4 \\ 
        \multicolumn{2}{r}{+ ICM} &  & 42.9~\increase{6.3} & 10.0~\decrease{1.6} & 13.4~\increase{2.0} \\
        \rowcolor{aliceblue!100} \multicolumn{2}{r}{+ ICM-r2} &  & 43.2~\increase{6.6} & 11.4~\decrease{0.2} & 14.5~\increase{3.1} \\
        \multicolumn{2}{l}{$\text{Doubao}^\dag$} & – & 67.1 & 67.3 & 43.8 \\ 
        \multicolumn{2}{r}{+ ICM} & & 70.1~\increase{3.0} & 68.4~\increase{1.1} & 46.9~\increase{3.1} \\
        \rowcolor{aliceblue!100} \multicolumn{2}{r}{+ ICM-r2} & & 71.6~\increase{4.5} & 68.0~\increase{0.7} & 47.9~\increase{4.1} \\
        \multicolumn{2}{l}{Qwen 2.5 VL} & 7B & 52.7 & 77.2 & 40.2 \\ 
        \multicolumn{2}{r}{+ ICM} & & 57.3~\increase{4.6} & 77.2~\nochange{} & 43.7~\increase{3.5} \\
        \rowcolor{aliceblue!100} \multicolumn{2}{r}{+ ICM-r2} & & 58.1~\increase{5.4} & 78.3~\increase{1.1} & 44.8~\increase{4.6} \\
        \multicolumn{2}{l}{$\text{UI-TARS 1.0}^*$} & 7B & 60.8 & 71.0 & 43.3 \\ 
        \multicolumn{2}{r}{+ ICM} & & 72.3~\increase{11.5} & 76.1~\increase{5.1} & 55.3~\increase{12.0} \\
        \rowcolor{aliceblue!100} \multicolumn{2}{r}{+ ICM-r2} & & 71.4~\increase{10.6} & 78.5~\increase{7.5} & 56.3~\increase{13.0} \\
        \multicolumn{2}{l}{$\text{UI-TARS 1.5}^*$} & 7B & 71.1 & 44.6 & 32.9 \\ 
        \multicolumn{2}{r}{+ ICM} & & 78.2~\increase{7.1} & 52.9~\increase{8.3} & 47.8~\increase{14.9} \\
        \rowcolor{aliceblue!100} \multicolumn{2}{r}{+ ICM-r2} & & 80.2~\increase{9.1} & 53.5~\increase{8.9} & 50.2~\increase{17.3} \\
        \noalign{\hrule height 1.0pt}
    \end{tabular}
\end{table}

\begin{table*}[!t]
    \renewcommand{\arraystretch}{1.15}  
    \scriptsize                      
    \centering
    \caption{\textbf{GUI grounding accuracy on ScreenSpotV2.} $^*$ indicates reproduced open-source agent performance. $\color{purple}{\uparrow}$ and $\color{gray!48}\downarrow$ respectively represent the performance changes relative to the base agents.} 
    \begin{tabular}{y{56}x{23}x{25}y{31}y{33.5}y{31}y{31}y{31}y{31}y{31}}
        \noalign{\hrule height 1.0pt}
        \hline
        \multirow{2}{*}{Model} & \multirow{2}{*}{Method} & \multirow{2}{*}{Size} & \multicolumn{2}{c}{Mobile} & \multicolumn{2}{c}{Desktop} & \multicolumn{2}{c}{Web} & \multirow{2}{*}{Avg.}\\ \cmidrule(lr){4-5} \cmidrule(lr){6-7} \cmidrule(lr){8-9}
                             & & & Text & Icon & Text & Icon & Text & Icon & \\ \hline
        GPT-4o & ZS & – & 30.5 & 23.2 & 20.6 & 19.4 & 11.1 & 7.8 & 18.8 \\
        OS-Atlas-Base & ZS & 7B & 93.0 & 72.9 & 91.8 & 62.9 & 90.9 & 74.3 & 82.5 \\
        SeeClick & SFT & 9.6B & 78.0 & 52.0 & 72.2 & 30.0 & 55.7 & 32.5 & 53.4 \\
        Aguvis & SFT & 7B & 95.6 & 77.7 & 93.8 & 67.1 & 88.3 & 75.2 & 84.4 \\
        \hline
        
        \multicolumn{2}{l}{Qwen 2.5 VL} & 7B & 84.8 & 59.7 & 72.1 & 52.1 & 69.2 & 46.3 & 65.0 \\ 
        \multicolumn{2}{r}{+ ICM} & & 87.9~\increase{3.1} & 70.1~\increase{10.4} & 79.4~\increase{7.3} & 57.1~\increase{5.0} & 74.7~\increase{5.5} & 49.2~\increase{2.9} & 70.4~\increase{5.4} \\
        \rowcolor{aliceblue!100} \multicolumn{2}{r}{+ ICM-r2} & & 89.7~\increase{4.9} & 68.2~\increase{8.5} & 78.9~\increase{6.8} & 54.3~\increase{2.2} & 76.9~\increase{7.7} & 51.2~\increase{4.9} & 71.1~\increase{6.1} \\
        
        \multicolumn{2}{l}{$\text{UI-TARS 1.0}^*$} & 7B & 93.1 & 82.4 & 94.8 & 76.4 & 91.8 & 84.2 & 88.1 \\ 
        \multicolumn{2}{r}{+ ICM} & & 94.5~\increase{1.3} & 83.1~\increase{0.7} & 93.2~\decrease{1.6} & 77.9~\increase{1.5} & 93.5~\increase{2.0} & 84.7~\increase{0.5} & 88.7~\increase{0.6} \\
        \rowcolor{aliceblue!100} \multicolumn{2}{r}{+ ICM-r2} & & 94.2~\increase{1.1} & 83.7~\increase{1.3} & 95.7~\increase{0.9} & 78.7~\increase{2.3} & 92.1~\increase{0.3} & 84.7~\increase{0.5} & 89.0~\increase{0.9} \\
        
        \multicolumn{2}{l}{$\text{UI-TARS 1.5}^*$} & 7B & 96.2 & 84.3 & 94.3 & 84.2 & 94.4 & 86.6 & 90.8 \\ 
        \multicolumn{2}{r}{+ ICM} & & 96.5~\increase{0.3} & 85.3~\increase{1.0} & 95.4~\increase{1.1} & 84.3~\increase{0.1} & 94.2~\decrease{0.2} & 85.2~\decrease{1.4} & 90.2~\decrease{0.6} \\
        \rowcolor{aliceblue!100} \multicolumn{2}{r}{+ ICM-r2} & & 97.3~\increase{1.1} & 84.2~\decrease{0.1} & 92.4~\decrease{1.9} & 84.4~\increase{0.2} & 95.5~\increase{1.1} & 87.7~\increase{1.1} & 91.0~\increase{0.2} \\
        \noalign{\hrule height 1.0pt}
    \end{tabular}
    \label{tab_grounding}
\end{table*}

\textbf{Evaluation.}
To evaluate the performance of the agent after being guided by the critic model, we evaluate the agent's task understanding, grounding, and planning capabilities on the AndroidControl and GUI-Odyssey test sets. For grounding ability, we additionally evaluate performance on ScreenSpotV2~\cite{cheng2024seeclick}. According to the input, the settings on AndroidControl can be divided into low-level tasks and high-level tasks. High-level tasks only input the global instruction to the agent, while low-level tasks will additionally input the single-step action plan.

Importantly, our critic models are trained exclusively under the high-level setting, without access to single-step action plans. This setting is more challenging yet realistic, as it relies solely on global instructions and interaction history. For consistency, the critic operates under the same high-level setting even when evaluating low-level agents on AndroidControl. GUI-Odyssey follows the high-level setting by default, and both the agent and critic are evaluated accordingly.



\textbf{Comparison.}
To verify the effectiveness of the proposed evaluation model in guiding existing agent models during testing, we selected a wide range of models. Among the closed-source models, we selected GPT4o~\cite{hurst2024gpt} and Doubao~(UITARS 1.5)~\cite{doubao}, and implemented action acquisition through API calls. For open source models, we reproduced Qwen 2.5 VL~\cite{bai2025qwen2}, UI-TARS 1.0~\cite{qin2025ui}, and UI-TARS 1.5~\cite{qin2025ui}, where Qwen 2.5 VL is a general multimodal understanding model and UI-TARS is a model fine-tuned for GUI agent tasks. We used the official prompt template to reproduce basic performance and test the superposition evaluation model. The original UI-TARS requires historical actions and up to 5 images of historical steps as input. To simplify the operation, we only let the model refer to the text description of the historical steps and discard the excessive historical image input. For detailed prompt and inference scripts, please refer to the supplementary material.

As a performance comparison, we selected Zero Shot~(ZS) model OS-Atlas-Base~\cite{wu2024atlas}, SFT-tuned SeeClick~\cite{cheng2024seeclick}, Aria-UI~\cite{yang2024aria}, and Aguvis~\cite{xu2024aguvis}, and RFT-tuned UI-R1~\cite{lu2025ui} and GUI-R1~\cite{xia2025gui}.

\textbf{Evaluation Metrics.}
For planning tasks, in line with OS-Atlas~\cite{wu2024atlas}, we report action type prediction accuracy~(Type), click point prediction accuracy~(GR), and step success rate~(SR). Specifically: \textbf{Type} measures the exact‐match accuracy between predicted and ground‑truth action types (e.g., \emph{Click} vs. \emph{Swipe}). \textbf{GR} evaluates grounding performance via click point prediction accuracy in specific action types (e.g., \emph{Click}). \textbf{SR} is the step‑wise success rate: a step is counted as successful only if both the predicted action and its associated arguments (e.g., click coordinates or input text) match the ground truth. For grounding tasks, we use click point prediction accuracy as our evaluation metric.

\begin{figure}[t]
    \centering 
    \begin{subfigure}[t]{0.45\columnwidth}
        \centering
        \includegraphics[width=\textwidth]{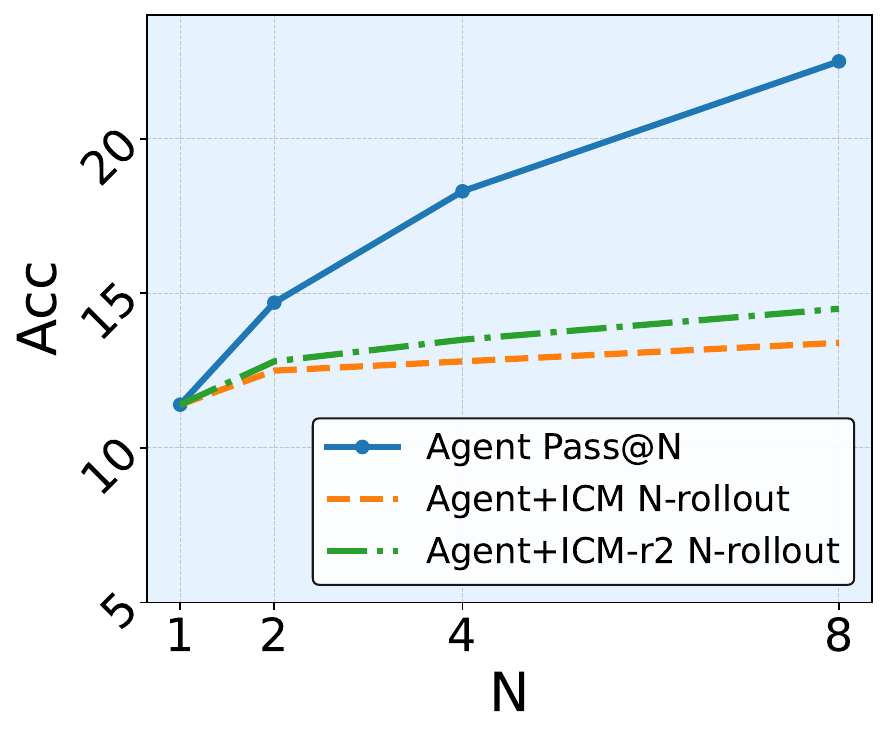}
        \captionsetup{width=.9\linewidth}
        \caption{GPT4o on GUI-Odyssey}
        \label{fig:sub1}
    \end{subfigure}
    \begin{subfigure}[t]{0.45\columnwidth}
        \centering
        \includegraphics[width=\textwidth]{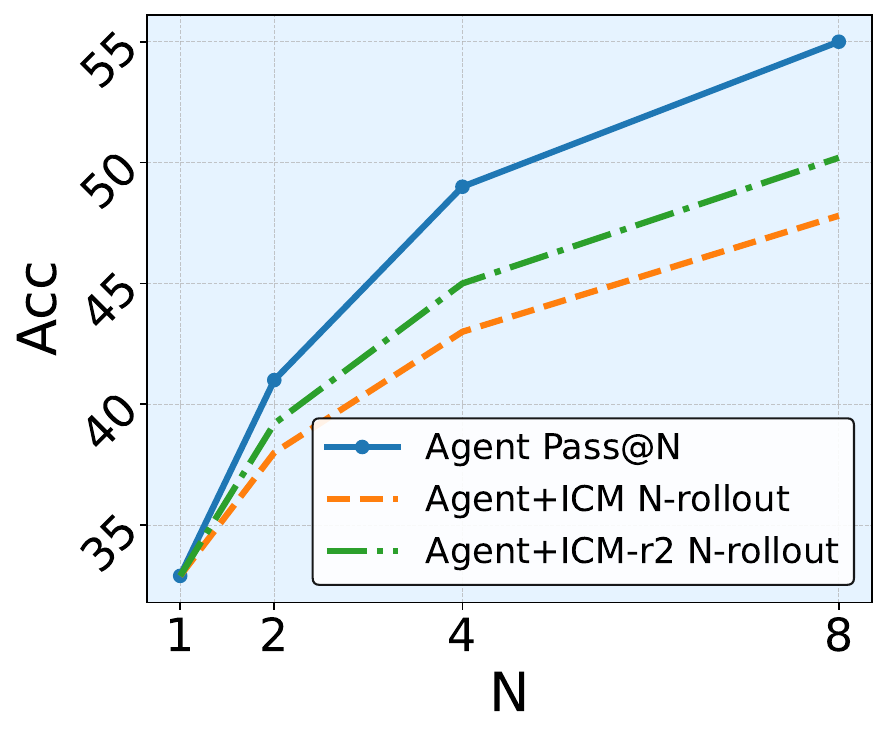}
        \captionsetup{width=.9\linewidth}
        \caption{UI-TARS 1.5$^*$ on GUI-Odyssey}
        \label{fig:sub2}
    \end{subfigure}

    \caption{\textbf{Performance improvements of Pass@N and $N$-rollout.}}
    \label{fig_N}
\end{figure}

\subsection{Experimental Results}
As shown in Table~\ref{tab_plan_android} and~\ref{tab_plan_odyssey}, the proposed ICM and ICM-r2 achieve extensive performance improvements for both zero-shot and fine-tuned GUI agents. On the AndroidControl-High test, ICM can improve agents' SR performance by up to $9.3\%$, while ICM-r2 can further improve it by an average of $1.32\%$. The same trend is also observed on GUI-Odyssey, demonstrating that existing models have the potential to correctly infer actions. ICM can leverage this potential to effectively improve existing agents during testing, and data flywheels can further amplify these improvements. For agents with lower basic performance, ICM can significantly improve their performance to an advanced level, which also demonstrates the potential of the model itself and the stimulation ability of the critic model.
In terms of generalization, GPT4o, Doubao, and Qwen 2.5 VL were not included in the construction of the GAIA $\mathcal{D}$ and $\mathcal{D}^+$, but ICM and ICM-r2 still achieved significant performance improvements, demonstrating the inherent consistency of real action space data. The real action sampling in the proposed data flywheel effectively covers this space, providing effective support for critic training.

Both our ICM and ICM-r2 use a high-level approach to judge correctness and guide action, meaning the critic model is not aware of the current action plan. This setting is more consistent with practical applications, where only global instructions are given, and the agent must independently reason about each step's plan and action. For AndroidControl-Low, the agent is aware of the current action plan, resulting in higher baseline performance. Despite this, our ICM still achieves a certain degree of performance improvement, demonstrating the effectiveness of our proposed approach.

Table~\ref{tab_grounding} shows the improvement of ICM and ICM-r2 on the grounding ability of agents on ScreenSpotv2. The ScreenSpotv2 data is not included in the proposed GAIA, and as a single-step operation, its environmental information is not completely consistent with the aforementioned datasets. Even so, our evaluation model still improves the performance of agents, which is sufficient to prove the validity of the data flywheel definition.

\subsection{Ablation Study}
While utilizing ICM and ICM-r2, we used $N$-rollout to improve the test-time performance of existing GUI agents, where $N$ is set to 8 by default. To measure the impact of $N$ on final performance, we selected GPT4o and UI-TARS 1.5$^*$ as representative closed-source and open-source models, respectively, and compared their SR at different $N$ values on GUI-Odyssey. We also measured the models' Pass@N during the rollout process to reflect the model's performance ceiling. As shown in Figure~\ref{fig_N}, the increase in Pass@N accuracy reveals the potential of the agents themselves, while the evaluation model approaches this upper limit through N-rollout. The improvement in ICM-r2 and the gap between the upper limit provide potential performance gains for further cycles of GAIA.

\subsection{Computation Efficiency}

\begin{table}[t]
    \centering
    \small
    \begin{minipage}[t]{0.48\textwidth}
        \renewcommand{\arraystretch}{1.15} 
        \setlength{\tabcolsep}{4pt}      
        \caption{\textbf{The relationship between computation cost and the number of $N$.} Time is measured in seconds (s).}
        \label{tab_cost_analysis}
        \begin{tabular}{y{30}x{23}x{23}x{23}x{23}} 
            \noalign{\hrule height 1.0pt}
            \hline
            \multirow{2}{*}{Metric} & \multicolumn{4}{c}{$N$ } \\ \cmidrule(lr){2-5}
             & 1 & 2 & 4 & 8 \\
            \hline
            Actor  & 1.0143 & 1.0874 & 1.1316 & 1.2617 \\
            Critic & 0.4739 & 0.6746 & 1.1018 & 1.9897 \\
            \rowcolor{aliceblue!100} Total & 1.4882 & 1.7620 & 2.2334 & 3.2514 \\
            \noalign{\hrule height 1.0pt}
        \end{tabular}
    \end{minipage}
    \hfill 
    \begin{minipage}[t]{0.48\textwidth}
        \renewcommand{\arraystretch}{1.15} 
        \centering
        \caption{\textbf{Critic comparison.} The performance is calculated as (Qwen2.5 VL 7B w/ critic) - (Qwen2.5 VL 7B w/o critic).}
        \label{tab_reward}
        \begin{tabular}{lcccc}
            \noalign{\hrule height 1.0pt}
            \hline
            \multirow{2}{*}{Model} & \multirow{2}{*}{$N$} & \multicolumn{3}{c}{AndroidControl-High} \\ \cmidrule(lr){3-5}
            & & $\Delta$ Type & $\Delta$ GR & $\Delta$ SR \\
            \hline
            UI-Genie-RM & 10 & – & – & 0.3 \\
            \rowcolor{aliceblue!100} ICM & 8 & 1.0 & 5.0 & 2.8 \\
            \rowcolor{aliceblue!100} ICM-r2 & 8 & 1.4 & 5.0 & 3.2 \\
            \noalign{\hrule height 1.0pt}
        \end{tabular}
    \end{minipage}
\end{table}

\begin{table}[t]
    \renewcommand{\arraystretch}{1.15} 
    \centering
    \caption{\textbf{Impact of differences in critic model attributes on accuracy and guidance.} The tests are conducted on the GUI-Odyssey dataset using UI-TARS 1.5$^*$ as the base model.}
    \label{tab_critic}
    \begin{tabular}{y{60}x{50}x{40}x{40}x{40}}
        \noalign{\hrule height 1.0pt}
        \hline
        \multirow{2}{*}{Model} & \multirow{2}{*}{Critic Acc} & \multicolumn{3}{c}{GUI-Odyssey} \\ \cmidrule(lr){3-5}
         & & Type & GR & SR \\
        \hline
        UI-TARS 1.5$^*$ & – & 71.1 & 44.6 & 32.9 \\
        \multicolumn{1}{r}{+ RCM} & 70.82\% & 75.6 & 49.2 & 44.1 \\
        \rowcolor{aliceblue!100} \multicolumn{1}{r}{+ ICM} & 83.19\% & 78.2 & 52.9 & 47.8 \\
        \rowcolor{aliceblue!100} \multicolumn{1}{r}{+ ICM-r2} & 83.56\% & 80.2 & 53.5 & 50.2 \\ 
        \noalign{\hrule height 1.0pt}
    \end{tabular}
\end{table}

We conduct an efficiency analysis to verify that the critic-guided TTS does not introduce prohibitive computational overhead.
For open-source models, we leverage vLLM for actor inference and multi-GPU parallelism for critic scoring. As shown in Table~\ref{tab_cost_analysis}, the total latency at $N{=}8$ is only approximately $2.2\times$ that of $N{=}1$, indicating that the time complexity scales sub-linearly with $N$ due to effective parallel execution. Notably, the dominant cost comes from base-agent rollouts rather than critic inference, as the critic operates as a lightweight binary evaluator with batched evaluation over $N$ candidates.

For closed-source models, since API calls are inherently parallelizable, varying $N$ from 1 to 8 introduces negligible additional latency, which is also adopted by existing GUI agent frameworks such as GTA1~\cite{yang2025gta1}. In practice, a single rollout typically involves 1,000 -- 2,000 input tokens and about 30 output tokens. The corresponding API cost can be estimated based on the pricing of the deployed model.
These results confirm that the proposed method can be deployed with acceptable overhead in practical settings.

\begin{figure*}[t]
    \centering
    \includegraphics[width=0.85\textwidth]{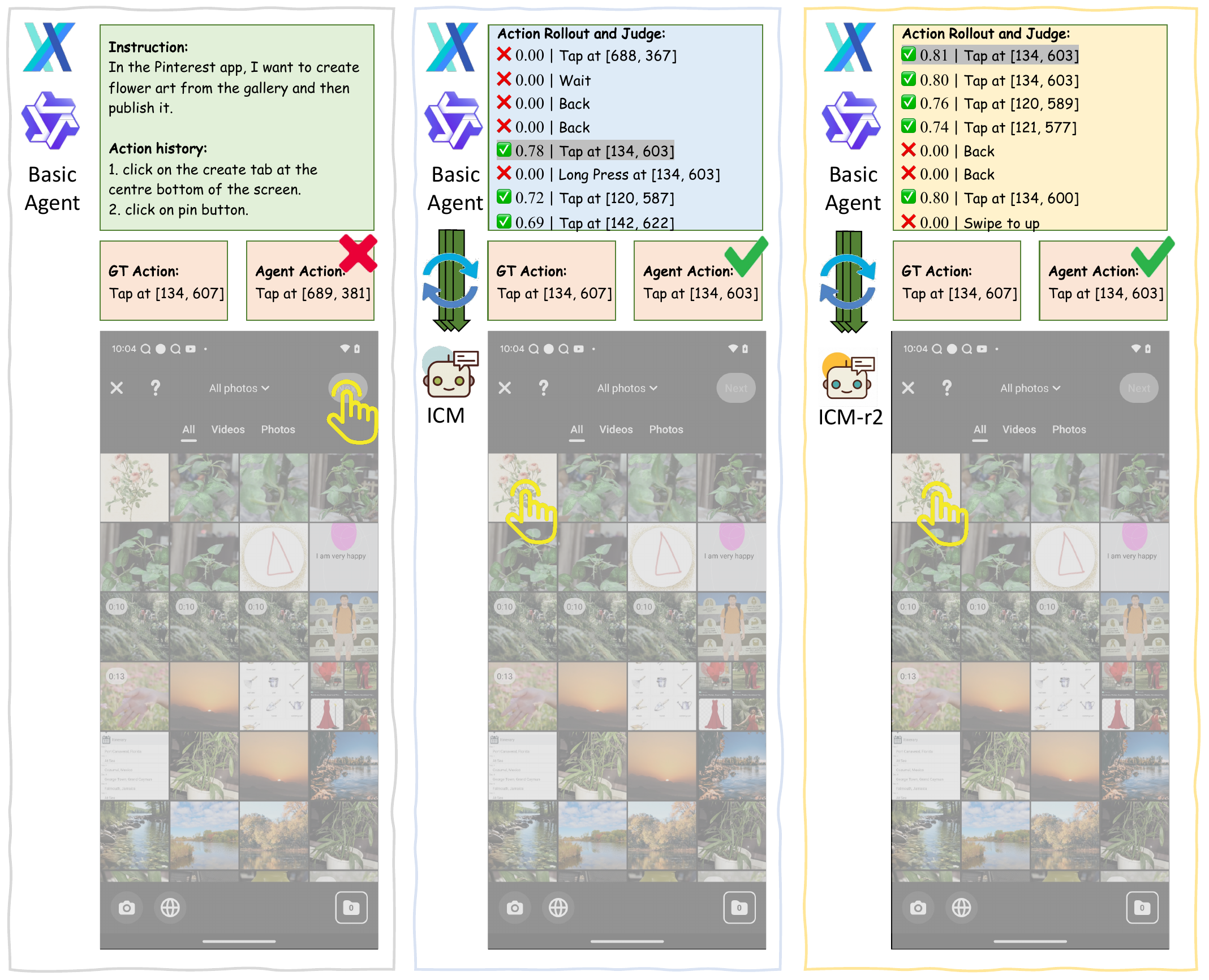}
    \caption{\textbf{Visualization result.} The basic agent selects the wrong action. Based on the action rollout, both ICM and ICM-r2 select the correct action from the candidates.}
    \label{vis_f1}
\end{figure*}

\subsection{Qualitative Experiment}
\textbf{Critic Model Comparison.}
To evaluate the effectiveness of the proposed discriminant model, we compared the accuracy of the Qwen 2.5 VL using a best-of-N approach to guide inference on AndroidControl-High with UI-Genie-RM~\cite{xiao2025ui}. As shown in Table~\ref{tab_reward}, due to the use of real action data, the proposed ICM significantly improves the accuracy of the base model, and ICM-r2 can further expand the advantage.

\textbf{Intuitive and Reasoning Critic.}
To verify that the intuitive judgment proposed in this article is superior, this section implements a critic model based on reinforcement learning design. Specifically, the input of the Reasoning Critic Model~(RCM) is consistent with ICM, and the output includes  \xmltag{thinking}...\allowbreak\xmltag{/thinking} and \xmltag{critic}...\xmltag{/critic}, which are supervised by format reward and critic reward. The training of RCM is achieved through Group Relative Policy Optimization~(GRPO). Considering the property of reinforcement learning, which is that it can stimulate model capabilities with less data, we randomly sampled 30k data from $\mathcal{D}^+$ to train RCM. This data includes samples from two rounds of GAIA and has the same distribution as the training data for ICM-r2. To intuitively compare the discriminative performance of different critic models, we collected the GAIA test set in a high-level manner on the AndroidControl and GUI-Odyssey test sets in the same way as we collected the training data.

\begin{figure*}[t]
    \centering
    \includegraphics[width=0.78\textwidth]{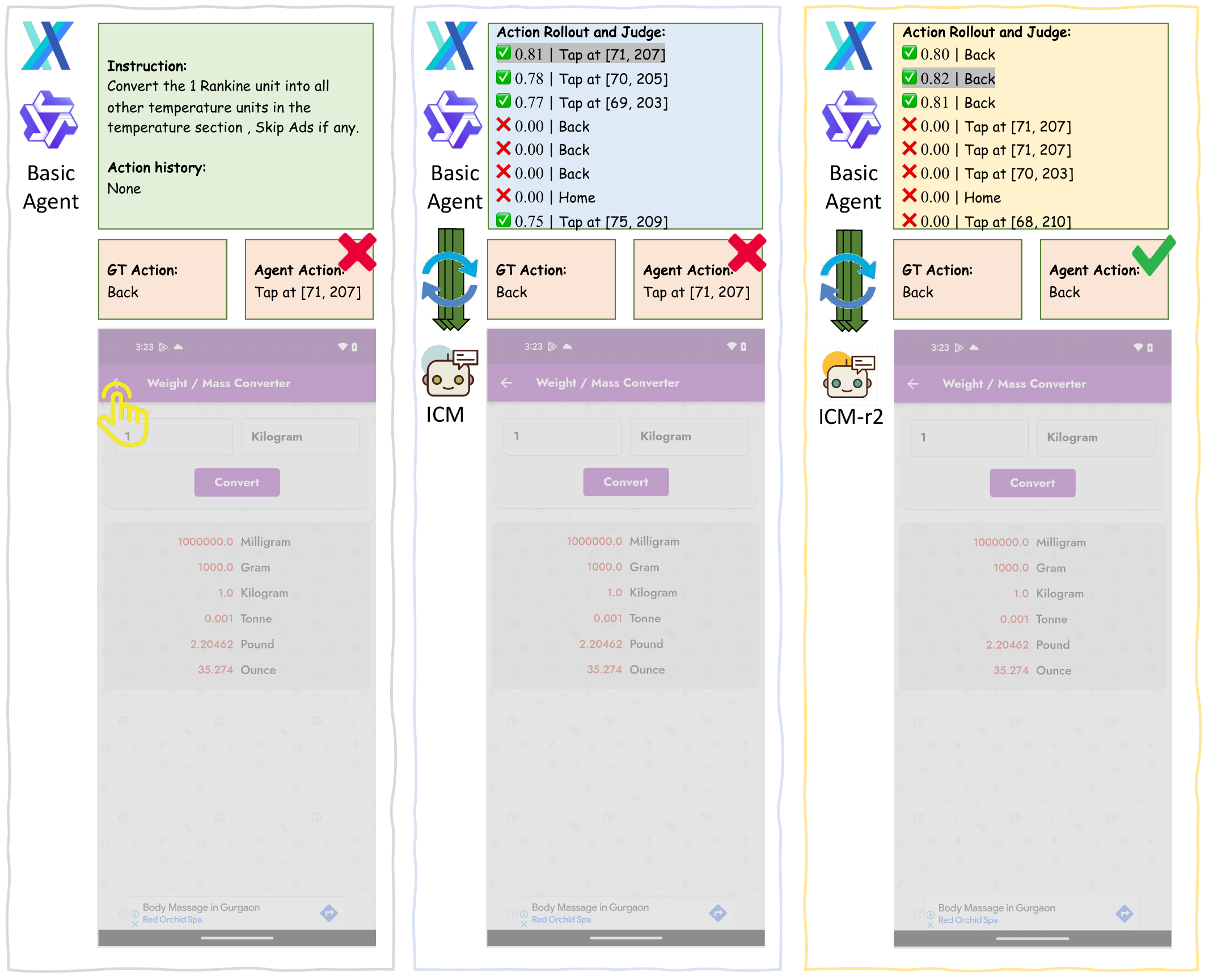}
    \caption{\textbf{Visualization result.} ICM fails to select the correct one from the rollout candidates, while the enhanced ICM-r2 guides the correct selection.}
    \label{vis_f3}
\end{figure*}

Table~\ref{tab_critic} shows that the proposed ICM achieves an accuracy of 83.19\% for correctness assessment, providing a foundation for action guidance. ICM-r2, benefiting from improved data quality, further achieves an accuracy of 83.56\%. In contrast, RCM's classification accuracy is 70.82\%, indicating that the thinking component fails to significantly contribute to the final assessment. In terms of action guidance accuracy, while UI-TARS 1.5$^*$ under RCM guidance outperforms the original model, it still falls short of ICM. This experimental result demonstrates that intuitive judgment outperforms reasoning for improving agents using a critic model.

Please refer to the supplementary material for more experimental results.

\subsection{Case Study}
To provide an intuitive understanding of the critic-guided TTS, we present qualitative examples in Figure~\ref{vis_f1} and Figure~\ref{vis_f3}. In Figure~\ref{vis_f1}, the basic agent selects an incorrect action, while both ICM and ICM-r2 successfully identify the correct one from the rollout candidates. Figure~\ref{vis_f3} illustrates a more challenging case where ICM fails to correct the erroneous action, yet ICM-r2 still guides the agent toward the correct selection. These cases demonstrate not only the effectiveness of the proposed critic mechanism but also the improved robustness of ICM-r2 in handling difficult scenarios. 

However, the failure of ICM in Figure~\ref{vis_f3} does not stem from an inherently incorrect action, but from the inability to distinguish between semantically equivalent yet differently defined actions under the benchmark (clicking a Back icon vs.\ triggering the system Back action). This corresponds to a multi-path scenario, where multiple valid actions achieve the same goal, suggesting a direction for improving the data flywheel. More visualizations are available in the supplementary material.

\section{Conclusion}
In this work, we addressed the critical challenge of irreversible errors in GUI agents by proposing GAIA, a framework that unleashes their latent potential at test time. GAIA comprises a data flywheel that iteratively curates realistic action samples and the Intuitive Critic Model~(ICM) that evaluates action correctness, establishing a self-evolutionary cycle where enriched data trains an increasingly powerful critic~(ICM-r2). By leveraging a Best-of-N strategy, ICM enables agents to select more reliable actions without resource-intensive retraining. Experiments on both closed-source and open-source agents demonstrate significant performance gains, presenting a scalable solution for building more robust GUI agents.
In future work, we plan to unify high-level and low-level guidance methods and collect richer data through online testing to continuously iterate the data flywheel.

\bibliographystyle{splncs04}
\bibliography{main}
\end{document}